\pdfoutput=1

\documentclass[dvipsnames,11pt]{article}

\usepackage[preprint]{acl}

\usepackage{times}
\usepackage{latexsym}
\usepackage[T1]{fontenc}
\usepackage[utf8]{inputenc}
\usepackage{microtype}
\usepackage{inconsolata}
\usepackage{amsmath}
\usepackage{amssymb}
\usepackage{amsthm}
\usepackage{graphicx}
\usepackage{mathtools}
\usepackage{listings}
\usepackage{tikz}
\usepackage{siunitx}
\usepackage{tipa}
\usepackage{float}
\usepackage{bbm}
\usepackage{breqn}
\usepackage{tikz-dependency}
\usepackage{booktabs}
\usepackage{hyperref}
\usepackage{linguex}
\usepackage{multirow}
\usepackage{seqsplit}
\usepackage[shortlabels]{enumitem}
\usepackage{utils}

\usepackage{xcolor}

\usepackage{caption}
\usepackage{subcaption}
\usepackage{linguex}
\usepackage[ruled,vlined]{algorithm2e}

\definecolor{babypink}{rgb}{0.96, 0.76, 0.76}

\title{Cross-Modal Taxonomic Generalization in (Vision-) Language Models}

\author{Tianyang Xu,$^1$ Marcelo Sandoval-Casta\~neda,$^1$\\\bf Karen Livescu,$^1$ Greg Shakhnarovich,$^1$ Kanishka Misra$^2$\\$^1$Toyota Technological Institute at Chicago, $^2$The University of Texas at Austin\\\texttt{\{sallyxu, marcelo, klivescu, gregory\}@ttic.edu, kmisra@utexas.edu}}

\begin{document}
\maketitle
\begin{abstract}
What is the interplay between semantic representations learned by language models (LM) from \textit{surface form} alone to those learned from more grounded evidence? 
We study this question for a scenario where part of the input comes from a different modality---in our case, in a vision-language model (VLM), where a pretrained LM is aligned with a pretrained image encoder. 
As a case study, we focus on the task of predicting hypernyms of objects represented in images. 
We do so in a VLM setup where the image encoder and LM are kept frozen, and only the intermediate mappings are learned. We progressively deprive the VLM of explicit evidence for hypernyms, and test whether knowledge of hypernyms is recoverable from the LM. 
We find that the LMs we study can recover this knowledge and generalize even in the most extreme version of this experiment (when the model receives no evidence of a hypernym during training). Additional experiments suggest that this cross-modal taxonomic generalization persists under counterfactual image-label mappings only when the counterfactual data have high visual similarity within each category. 
Taken together, these findings suggest that cross-modal generalization in LMs arises as a result of both coherence in the extralinguistic input and knowledge derived from language cues.

\vspace{.11em}
\hspace{.5em}\includegraphics[width=1.1em,height=1.1em]{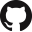}{\hspace{.5em}\parbox{\dimexpr\linewidth-2\fboxsep-2\fboxrule}{\small \url{https://github.com/sally-xu-42/cross-modal-taxonomic-gen}}}
\end{abstract}

\section{Introduction}
\label{sec:intro}

The widespread empirical success of language models (LMs) in showing sophisticated behavior has inspired a number of philosophical investigations of how learning the distribution of language \textit{form} contributes to the learning of \textit{meaning}. 
Many such discussions \citep{piantadosi2022meaning, pavlick2023symbols, mollo2023vector, mandelkern2024language} have converged into the broader viewpoint that LMs achieve a particular type of grounded meaning where the internal representation of a token is grounded in its \textit{relation} to the representations of other tokens.
This `relational grounding' viewpoint \citep{mollo2023vector} echoes a number of historical perspectives that emphasize the emergence of meaning from connections between symbols
\citep{harris1954distributional, firth1957synopsis, harman1982conceptual, harnad1990symbol, elman2004alternative}.

How does relational grounding from language interplay with other forms of meaning---e.g., those grounded in perceptual signals?
~For instance, LMs' knowledge of the concept \textit{bird} might include its syntax (that it is a noun), that it is a hypernym of other concepts (\textit{robin, sparrow}), and that it is thematically related to concepts like \textit{trees, branches, nests, flying}, etc. 
How does this knowledge impact LMs' behavior when they encounter a bird via a different modality---e.g., its visual representation? 

\begin{figure*}[!t]
    \centering
    \includegraphics[width=0.8 \linewidth]{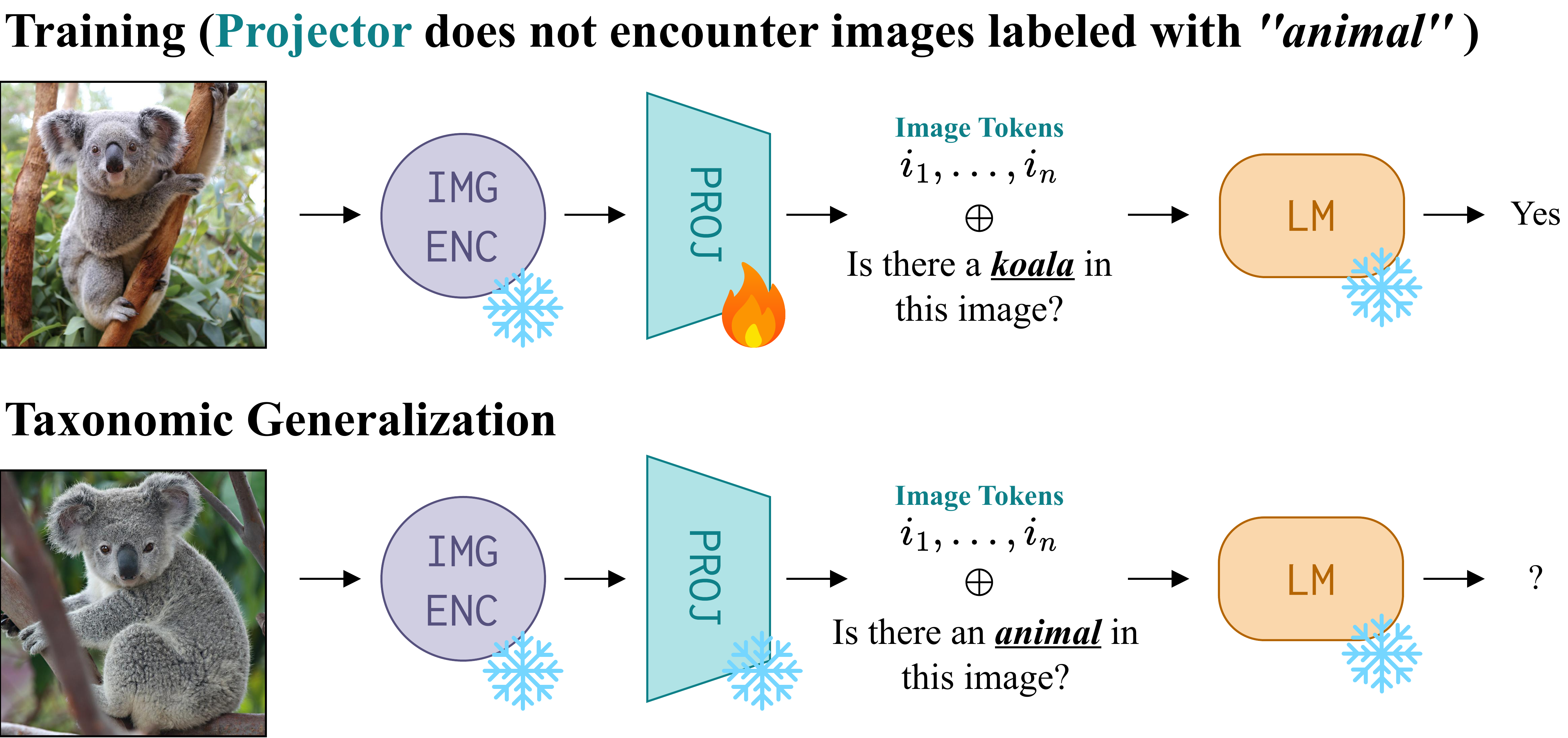}
    \caption{An instance of our experiments. During training, the projector is deprived of explicit supervision on high-level categories (\textit{hypernyms}, e.g., \textit{animal}) at various amounts, and is trained to detect the presence (and absence) of lower-level categories (e.g., \textit{koala}), keeping the image encoder and the LM backbone frozen. After training, the VLM is tested for generalization to hypernym categories, given previously unseen images.}
    \label{fig:fig1}
    \vspace{-1em}
\end{figure*}

This question can be answered by analyzing models that learn to map between the representations in LMs and non-textual modalities such as vision or speech. 
Modern vision-language models \citep[VLMs;][a.o.]{liu2023visual}---models that learn to map between pretrained image encoders and language models---are a particularly popular class of such models. 
Recent work has found evidence that the LM components of such models often overpower their image encoder counterparts, to the extent that the model sometimes predicts a response without relying on the visual input \citep{golovanevsky-etal-2025-pixels, lee-etal-2025-vlind, fu2025hidden, frank-allaway-2025-visage, hua2025vision}. 
While this has been interpreted as a bug, especially in perception-heavy tasks \citep{fu2024blink}, it also highlights the extent to which certain kinds of knowledge can be acquired from language alone.\footnote{For instance, predicting that a dog has four legs given an image of a dog with three legs \citep{frank-allaway-2025-visage} is indeed an issue, but can be attributed to the effect of \textit{generic} knowledge \citep[e.g., \textit{Birds can fly}, \textit{Lions have manes};][]{leslie2008generics} that is highly prevalent in language data \citep{gelman2010effects}.}

In this paper, we place particular emphasis on the latter takeaway, and take it a step further. 
We do so by systematically holding out a particular piece of knowledge during the training of VLMs, and test if it is recoverable directly from the language model's learned representations. 
Doing so allows us to answer the broader question of how knowledge learned \textit{exclusively} from language extends beyond it and interacts with the input signals from an extra-linguistic modality (here, images).
As a case study, we specifically focus on the ability of the models to perform \textbf{taxonomic generalization}: predicting the presence of a hypernym (higher-order category, e.g., \textit{bird}, \textit{animal}) in an input image, given that they have learned to identify \textit{only} lower-level categories (e.g., \textit{parrot, koala}) during training.
To what extent does the taxonomic knowledge learned by LMs generalize \textit{across modalities}?

Taxonomic knowledge is a worthwhile testing ground for our research question. 
Taxonomic organization is closely connected to categorization, a fundamental capacity of human cognition \citep{murphy2004big}---learning an entity's taxonomic category allows us to make systematic, default inferences (e.g., if the entity is a bird, then it likely has wings).
There have been a number of hypotheses about the learnability of category knowledge from linguistic cues, for both computational models \citep{hearst-1992-automatic, geffet2005distributional} and humans \citep{waxman1995words, wilson2023conceptual}.  Our work contributes to the study of how such category knowledge arises in computational models.

\citet{hearst-1992-automatic} proposed a set of distributional patterns that strongly signal taxonomic relations---e.g., patterns like ``\textit{Xs such as Ys}'' can serve as a cue to the inference that Y is a type of X.
Similarly, the distributional inclusion hypothesis \citep{geffet2005distributional} states that hyponyms tend to appear in a subset of the distributional contexts in which the hypernyms can appear---e.g., \textit{animal} can appear in all the contexts in which \textit{dog} can appear and additionally those where it does not. 
Therefore, it is conceivable that LMs, being powerful models of distributional semantics, might pick up on these cues to learn the relations between hypernym categories and their members.
More generally, language has been posited to be ``particularly influential when the information it expresses is not easily derived through first-hand observation and experience'' \citep{cimpian2008preschool}.
Our analyses contribute towards in-principle tests of the broader claims of these hypotheses (that taxonomic knowledge is retrievable from language data), in the extreme condition where the source of the target knowledge is restricted to 
language alone and the input comes from a non-language modality (here, vision).
So, if a learner (in this case, LMs) has made no explicit ``first-hand'' perceptual observations of \textit{any} hypernym category, to what extent can the knowledge it has acquired from language support the visual identification of such categories?

Our experiments (depicted in~\Cref{fig:fig1}) focus on a class of VLMs that are trained by learning a `projector' that maps image representations from a frozen image encoder to the embedding layer of a frozen pre-trained language model, similarly to \citet{liu2023visual} and \citet{merullo2023linearly}. 
We first train these models to answer questions that ask whether a given object-centric image contains a category (e.g., \textit{Is there a \textbf{cat} in this image?}). 
We then test whether they can obtain above-chance performance at identifying both leaf-level \textit{and} hypernym categories on images that do not appear in training. 
Having established that they do, we then test whether they can generalize their knowledge of hypernyms \textit{without} explicit vision-language supervision---i.e., only from their pretrained language representations. We do so by training additional instances of the projector while holding out various amounts of hypernyms. This experimental setup is inspired by recent work on controlled rearing or filtered-corpus training of LMs \citep{jumelet-etal-2021-language, misra-mahowald-2024-language, patil2024filtered, leong2024testing, yao2025both}, where LMs are trained on manipulated versions of input corpora and tested on their ability to generalize. 
In the most extreme version of this experiment, where there is \textbf{no evidence of a single hypernym} during the projector's training, we find that VLMs still achieve above-chance performance. 
That is, given an image of a \textit{sparrow}, VLMs are able to predict the presence of a \textit{bird}, without ever encountering image-question pairs containing the word ``bird''. 
This suggests that taxonomic knowledge is represented within LMs in a manner that can extend beyond modalities---i.e., \textbf{they can perform cross-modal taxonomic generalization}.

What is the role of the input signal in facilitating this generalization? 
Do LMs generalize arbitrarily, in a rule-like manner? 
For instance, if the projector were to be trained \textit{counterfactually} to predict ``cardinal'' when it was an image of \textit{hummus} and ``parrot'' when it was an image of a \textit{bandage}, would it still predict that these images contain \textit{birds} (\texttt{IF crow THEN bird})? This is not borne out in our findings. Models perform at chance under conditions where lexical items within a hypernym category are paired with randomly shuffled sets of images. Instead, they only generalize in counterfactual conditions that preserve the visual coherence of categories---i.e., when we shuffle images and leaf-level mapping \textit{within} the same category (e.g., \textit{cardinal} mapped to images of \textit{eagles} and \textit{parrot} mapped to images of \textit{pheasants}). 
This finding gives us preliminary insight about the \textbf{importance of input coherence} \citep{murphy1985role} \textbf{in order for cross-modal generalizations to emerge}, paving the way for future investigations across multiple modalities and phenomena.

\section{Related Work}
\label{sec:related}

Taxonomic knowledge (or hypernymy) is one of the most well-studied lexical relations in computational linguistics \citep{murphy2003semantic}, and has long been hypothesized to be extractable from naturally occurring language \citep{hearst-1992-automatic, geffet2005distributional, roller-erk-2016-relations}. 
This has been especially apparent in recent investigations of hypernymy in pretrained LMs \citep[][\textit{a.o.}]{hanna-marecek-2021-analyzing, moskvoretskii-etal-2024-large, regneri-etal-2024-detecting}. While hypernymy is also captured by VLMs \citep{qin2025visionandlanguage}, the extent to which it can \textit{possibly} emerge from their LM component alone is unclear.
Our methods shed light on this, since we restrict the knowledge of hypernymy to the LM backbone, and use image encoders that have not been supervised using text data.

Our findings also relate to recent proposals about the convergence in the representations of models trained on different modalities or domains \citep{cao2021explanatory, sorscher2022neural, li-etal-2024-vision-language, huh2024plato, wu2025semantic}. For instance, according to the `Platonic Representation Hypothesis' \citep{huh2024plato}, the representations learned by unimodal models of different modalities are converging since they are statistical models of the same world.
We shed light on the implications of these proposals when two unimodal models (the vision encoder and the LM) are aligned. When this alignment happens under counterfactual scenarios (\Cref{sec:counterfactual}), where the incoming input signals are at odds with the conceptual organization of the two models---posited to be similar under the aforementioned proposals---we see a severe drop in taxonomic generalization. 
This underscores the importance of such representational convergence in the extreme setting where explicit information about the phenomenon (here, hypernymy) is restricted to a model from one modality.

\section{Data, Models, and Methods}
\label{sec:data}

\subsection{Stimuli and Measures} 
\label{sec:stimuli-design}
We conduct our experiments using the THINGS database of labeled, object-centric images~\citep{things2019}. Specifically, we use a 1,216-category and 17,336-image subset of THINGS made available by \citet{rodriguez-etal-2025-characterizing}, who also provide hypernym annotations. The leaf categories are grouped under 53 hypernyms, with each leaf category belonging to 1.67 hypernyms on average (max = 6, min = 1). 
Our task stimuli consist of English polar questions with the format ``\textit{Is there a \texttt{\{category\}} in this image?}'', where we substitute ``\texttt{\{category\}}'' with the target category.  
Each image in our experiments is associated with two types of questions: a positive question, asking for the presence of the category depicted in the image (expecting the answer `Yes'), and a negative question, asking for the presence of a different category (expecting the answer `No'). 
For negative questions, we substitute the leaf/hypernym category in the question by sampling uniformly from other leaf/hypernym categories in our set---we do this at the image level, which means that the same positive question is associated with multiple negative questions throughout our experiments. 
For example, as shown in \Cref{fig:fig1}, the same test image of a \textit{koala} is associated with ``\textit{Is there a koala in this image?}'' as well as ``\textit{Is there an animal in this image?}'' as its positive questions.
Since each leaf-level category is associated with multiple images, we split the set of images for each category into training (70\%), validation (5\%), and test (25\%) sets. 
Detailed statistics are available in \Cref{appsec:dataset-details}.
Because we conduct negative sampling at the image level, the resulting number of ``yes'' and ``no'' instances is not balanced within each leaf/hypernym---e.g., in 100\% ablation test set, the \textit{mammal} category has 332 and 149 ``yes'' and ``no'' questions, respectively. 
We therefore use macro F1 per category as our evaluation measure.

\subsection{Modeling Choices} 

Our models follow the general visually-conditioned LM architecture used by many recent studies \citep[][a.o]{liu2023visual, Qwen2-VL, chen2023pali3visionlanguagemodels}.
Specifically they include three components: 1) an \textbf{image encoder}, which takes in an image and produces a sequence of $N$ image feature vectors, each corresponding to an image patch, and is pre-trained using standard self-supervision objectives; 2) a vision-language \textbf{projector}, which is a multi-layer perceptron that transforms the image features into image ``tokens'', which are a sequence of vectors with the same dimensionality as the language model's embedding layer; and 3) an \textbf{LM backbone}, which is pre-trained on text data, and takes in a sequence of image tokens, followed by the task question, and (in our case) produces an answer---either ``yes'' or ``no''. 
During training, the image encoder and the LM backbone are kept frozen and only the projector is trained, similarly to \citet{merullo2023linearly}. We do this to preserve the information contained in the pre-trained components. 
Our training objective is next-word prediction, as in standard VLM fine-tuning---we minimize the negative log probability of the correct answer token (yes/no) given an image and a question, and compute the loss only at the answer position.

For our image encoder, we primarily experiment with DINOv2 \citep[large;][]{oquab2024dino} since it is trained using a self-supervised learning objective on images, and has never encountered text data (not even class labels). 
We do this to ensure that the image encoder representations are devoid of any text supervision, which may leak information about hypernyms even when we remove them during the training of the projector. 
Nevertheless, we compare performance obtained using DINOv2 to that obtained using SigLIP \citep{zhai2023sigmoid}, a CLIP-style encoder trained using a contrastive objective to map between images and text, in order to test whether the presence/absence of text during image encoder training has any effect on our models' taxonomic generalization. 

For our LM backbone, we primarily report results using publicly-released Qwen3 LMs \citep{yang2025qwen3} from HuggingFace \citep{wolf2020huggingfacestransformersstateoftheartnatural}, specifically ones with 0.6B and 1.7B parameters. 
We also report results using Llama 3.2 models \citep{grattafiori2024llama3herdmodels} in \cref{appsec:llama}. 
In addition, we compare VLMs using these pre-trained LMs to their randomly-initialized counterparts, in order to quantify the effect of text exposure on cross-modal taxonomic generalization. 
Details about our training hyperparameters are listed in \Cref{appsec:training-details}.

\subsection{Experiment Design}
\label{subsec:exp-design}

Since the goal of our experiments is to test whether hypernymy relations can be recovered by the LM backbone of our model, we always include all leaf categories and their training set images in the projector training data, and manipulate only the number of hypernym category labels shown during training. 
We perform two kinds of manipulations:

\paragraph{Random Hypernym Ablation}
We randomly remove mappings between leaf images and their hypernym labels from the training set.  By removing a mapping $(x,h)$ between leaf concept $x$ and hypernym $h$, we mean that we remove all image-question pairs with images of $x$ and questions about $h$.  For example, removing the mapping \textit{(parrot,bird)} corresponds to removing all images of parrots with the question ``Is this a bird?'' (as well as with corresponding negative questions like ``Is this a tool?''). 
As a result, a model sees only partial evidence for a given hypernym---e.g, it sees the label \textit{bird} for all images of \textit{crow}, but not for \textit{parrot}. 
For each of the 53 hypernym labels in the training set, we perform this ablation by removing 10\%, 30\%, 50\%, 70\%, 90\%, or 100\% of the total possible leaf-image sets.

\paragraph{Systematic Hypernym Ablation} 
We remove a specified number of hypernyms from the training data entirely. 
Here a model sees only a subset of the unique hypernyms---e.g., it might encounter \textit{animal} and \textit{vegetable}, but never encounter \textit{bird} and \textit{vehicle}. 
We perform this ablation by removing 10 (19\%), 20 (37\%), 30 (57\%), 40 (75\%), and all 53 (100\%) hypernyms from the positive training examples, leaving only their leaf-level labels. A removed hypernym may still appear as a negative category in questions paired with other hypernyms.

\begin{figure}[!t]
    \centering
    \includegraphics[width=\linewidth]{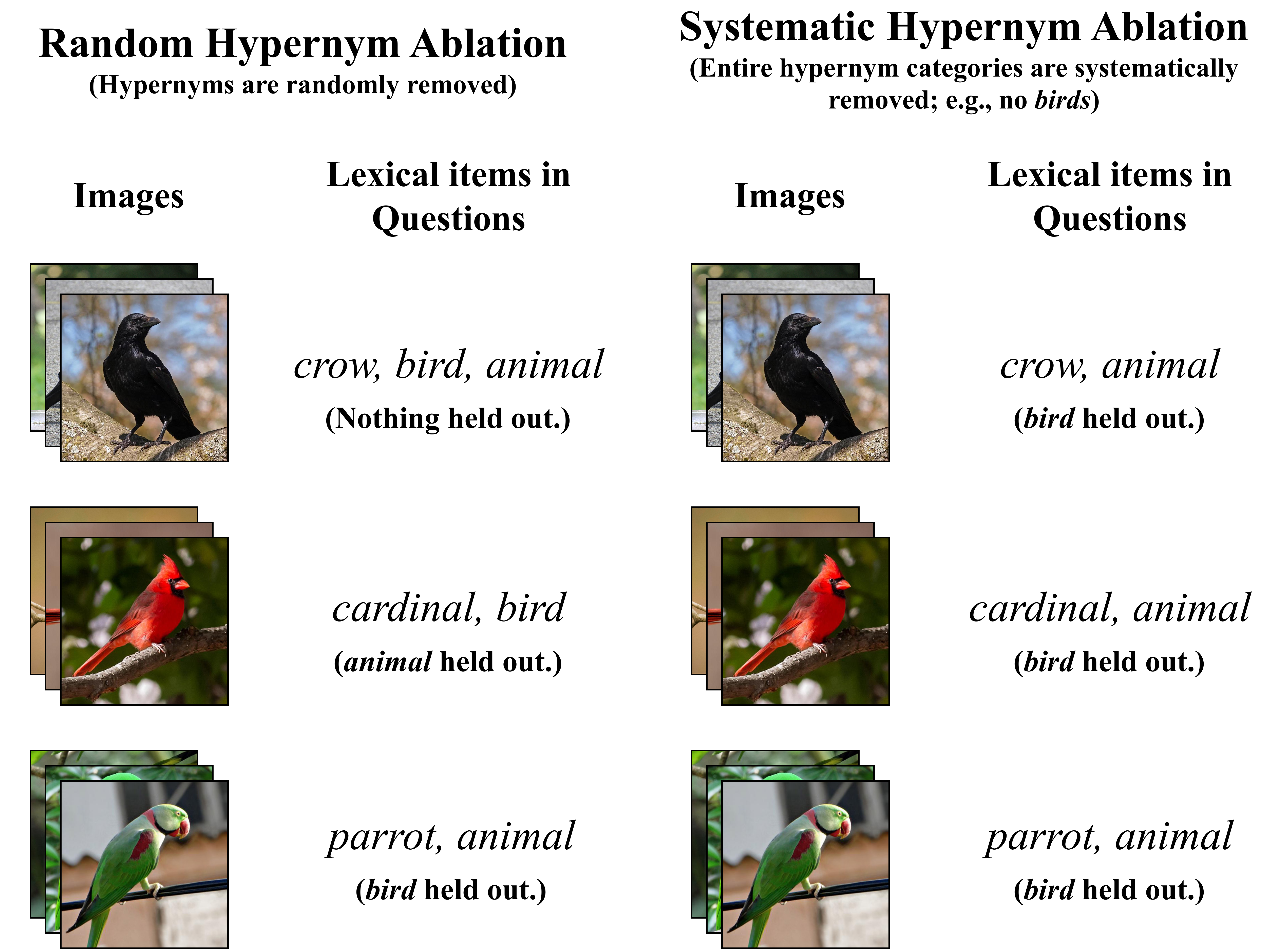}
    \caption{Depiction of our two ablations, for a sampled toy training set containing several images each of crows, cardinals, and parrots.}
    \label{fig:ablation}
    \vspace{-1em}
\end{figure}

\Cref{fig:ablation} shows examples of both ablations. 
Note that at 100\%, the two ablation types are equivalent. 
Appendix \ref{appsec:dataset-details} shows a detailed breakdown of our splits at various ablated percentages.

\section{Experiments}

\subsection{Preconditions to Generalization}
\label{sec:prelim-exp}

Before analyzing cross-modal taxonomic generalization, we first test the extent to which our model satisfies three preconditions. 
First, the LM backbone should have non-trivial knowledge about hypernymy---e.g., it should predict that \textit{animal} is a hypernym of \textit{dog, koala, sheep, giraffe,} etc. 
Next, the model should achieve non-trivial taxonomic generalization when the vision encoder has \textit{not} been informed using any text-based pretraining \citep[like in CLIP, see][]{radford2021learning}. 
Finally, the model should sufficiently learn the task of identifying categories in images, especially those that were held out from training. Below we report results from all three of these tests.

\paragraph{Hypernymy knowledge in LMs} 
Our first test focuses on the extent to which our LM backbones---namely, Qwen3-0.6B and Qwen3-1.7B---capture basic hypernymy relations. 
Arguably, this is the most critical precondition: LMs must possess the knowledge of the phenomenon we are trying to measure across modalities. 
We verify this for the set of categories in our dataset, by constructing text-only questions analogous to the subset of hypernym-focusing questions in our data (\Cref{sec:stimuli-design}). 
That is, for a question that asks ``\textit{Is there a/an \texttt{categoryX} in this image?}''~given an image of \texttt{categoryY}, we create an analogous question: ``\textit{Is it true that a/an \texttt{categoryY} is a type of \texttt{categoryX}?}'' This template is taken directly from \citet{qin2025visionandlanguage}, and results in a total of 16,903 questions (2,040 positive questions, and 14,863 negative questions). This imbalance is inherited from our main stimuli, as negative sampling was performed at the image level.
For each question, we compare the LM's probabilities for ``yes'' vs.~``no'', and evaluate using macro F1 scores. We find that both Qwen3 LMs achieve high F1 scores: 78.5 (0.6B) and 88.5 (1.7B), both of which are substantially greater than the F1 of a majority-label baseline (46.7).

\begin{figure}[!t]
    \centering
    \includegraphics[width=0.8\linewidth]{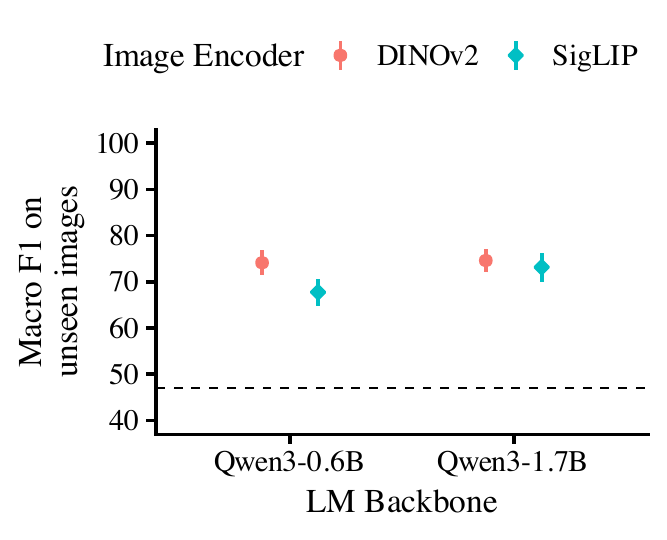}
    \caption{Macro F1s of VLMs for predicting the hypernym category for unseen images across LM backbones and image encoder types (DINOv2 vs.~SigLIP), in the experiment where all hypernyms are removed. Dashed line indicates majority-label baseline.}
    \label{fig:dino-siglip}
    \vspace{-1.5em}
\end{figure}

\begin{figure*}[!t]
    \centering
    \includegraphics[width=0.85\linewidth]{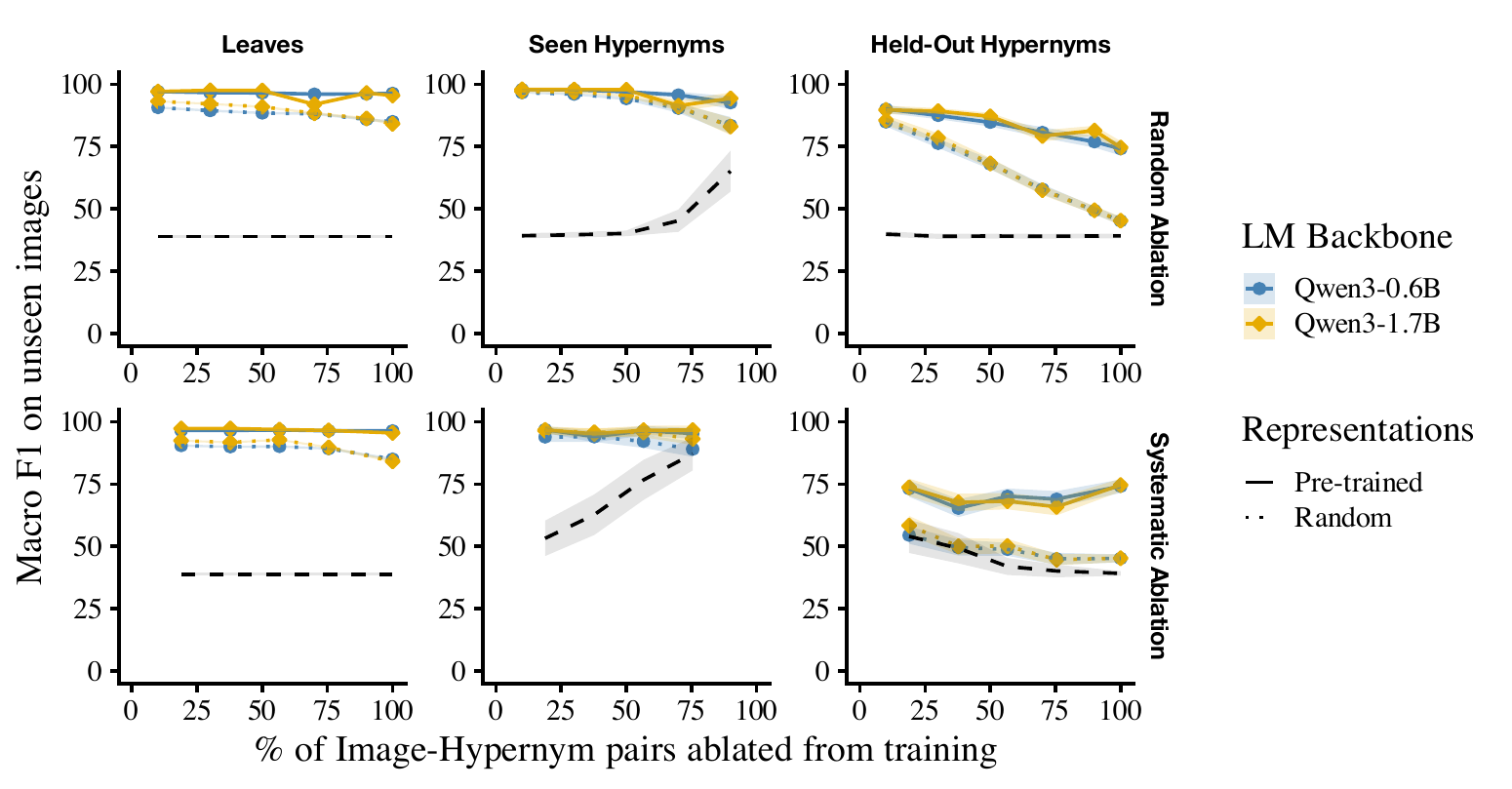}
    \caption{Macro F1 on unseen images across LM backbones, LM representations (pre-trained vs.~random), and hypernym ablation type (Random vs.~Systematic) at different amounts of exposure to hypernym categories, for various test splits. Dashed line indicates the majority-label baseline. Error bands represent 95\% confidence intervals across random seeds ($N$ = 3) and categories (hypernyms = 53, leaves = 1216).}
    \label{fig:exp1-main-results}
    \vspace{-1em}
\end{figure*}

\paragraph{Impact of text supervision in image encoders} 
A presupposition made by our research question is that it is only the LM backbone that is the source of textual/language exposure for the model.
The extent to which this presupposition holds depends on our choice of image encoder---while DINOv2 was trained with self-supervision \textit{only} on vision tasks (presupposition holds), SigLIP's loss function uses both text and image signals (presupposition does not hold). 
What is the impact of the image encoder's exposure to text on models' taxonomic generalization? 
To test this, we compare the macro F1s of models with fixed LM backbones, differing only in their image encoders (DINOv2 and SigLIP), on the specific sub-experiment where the model has not received \textit{any} exposure to a hypernym during VLM training (i.e., the strictest version of our experiments). 
\Cref{fig:dino-siglip} shows these results for the 0.6B and 1.7B versions of the Qwen3 LM. 
We use a linear mixed effects regression model to predict the macro F1 using image encoder and LM backbone as fixed effects, with random slopes for run seed and hypernym category. 
We do not find a significant effect of the image encoder ($t$ = 0.91, $p$ = 0.45). 
Since performance obtained using DINOv2 as our encoder is statistically equivalent to that obtained using SigLIP, we report all our subsequent results in the main text using DINOv2.

\vspace{-.05in}
\paragraph{Do our models learn the task?} 
Our final precondition is that the model learns to identify the categories in training in a generalizable manner---i.e., it does not simply memorize the training set and can identify leaf (and hypernym, when applicable) categories even when conditioned on images not seen during training. 
The two leftmost columns in \Cref{fig:exp1-main-results} (labeled ``Leaves'' and ``Seen Hypernyms'') show macro F1s of models for predicting the presence of leaf categories as well as hypernyms seen during training, on unseen images. 
We see that all models, including ones with randomly initialized LM backbone representations, achieve very high macro F1 scores at all levels of hypernym exposure, for all ablation types. 
That is, our models satisfy the precondition of learning the task.

\subsection{Cross-Modal Taxonomic Generalization}
\label{sec:exp1}

Having shown that the LM backbones capture hypernymy, that the trained VLMs can generalize to unseen images regardless of whether or not the image encoder receives textual supervision, and that they have learned to perform the task, we now turn to our main result, shown in the rightmost column of \Cref{fig:exp1-main-results}, named ``Held-Out Hypernyms''. 
We see that regardless of the amount of exposure to a hypernym category, LM backbones with pre-trained representations achieve consistently above-chance performance in predicting the presence of hypernyms in images not seen during the models' training. 
This is observed for both types of ablations, and even at 100\% ablation, where the model never sees \textit{any} hypernym label during training.
Models with randomly initialized LM backbone representations consistently underperform their pre-trained counterparts, and are close to chance in the absence of any hypernym-level supervision. 
These results also hold for Llama 3.2, which we report in \Cref{fig:main-exp-llama} in the Appendix.
These results suggest that the pre-trained representations of LMs \textit{can} facilitate cross-modal taxonomic generalization.
We analyze how this generalization varies across depths of taxonomic organization in \Cref{app:depthwise}.

\subsection{On the Arbitrariness of Cross-Modal Taxonomic Generalization}
\label{sec:counterfactual}

The results thus far show that LM representations can facilitate cross-modal taxonomic generalization---i.e., our models can predict that an image of a crow contains a bird, even though they have never encountered an instance of an image being labeled with ``\textit{bird}''. 
How abitrary is this behavior? 
Do LMs simply execute the rule, ``\texttt{IF crow THEN bird?}'' regardless of what the image-\textit{crow} mapping is---e.g., would they still predict that there are birds in images of \textit{kayak} and \textit{hummus}, if during training they encountered images of a \textit{kayak} labeled with \textit{crow}, and \textit{hummus} with \textit{cardinal}?
Or, do LM representations expect members of a category to \textit{cohere} in systematic ways \citep{murphy1985role}? 
For instance, members of the \textit{bird} category are likely to share a multitude of features---\textit{beaks}, \textit{wings}, \textit{feathers}, etc.---and this could foster the development of a cohesive category in the visual representations of its members, which could be the key to facilitating taxonomic category-based generalization within the model.

\begin{figure}[!t]
    \centering
    \includegraphics[width=0.8\linewidth]{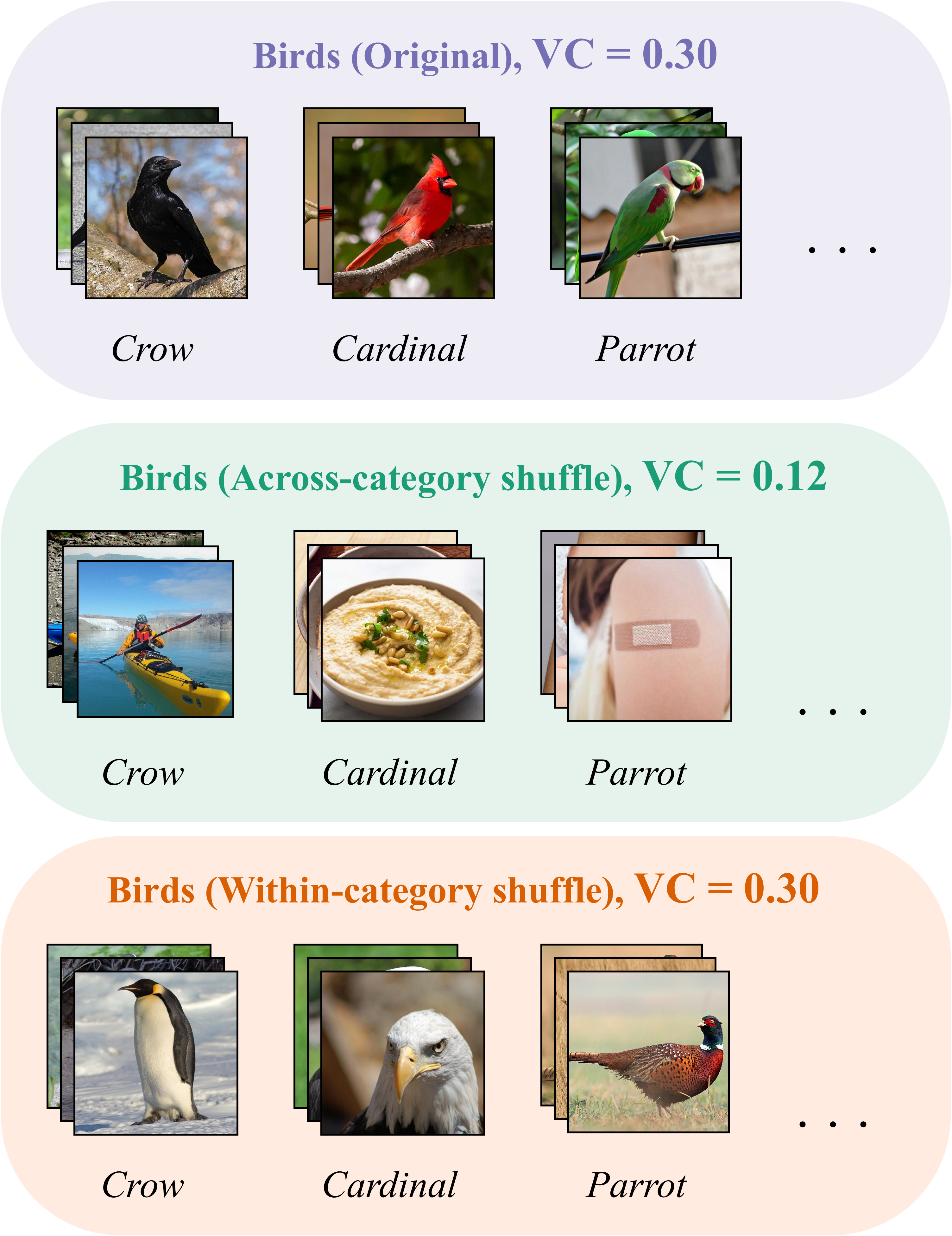}
    \caption{Examples of image-leaf mappings resulting from our counterfactual shuffles, in comparison with the original configuration (top). VC indicates the visual coherence of the category under the data configuration.}
    \label{fig:counterfactual-data}
    \vspace{-1em}
\end{figure}

To adjudicate between these possibilities, we repeat our experiments with two counterfactual versions of our training sets, each aiming to isolate a possible mechanism (arbitrary, rule-based taxonomic generalization, vs.~taxonomic generalization facilitated by visual coherence). 
We create these counterfactual datasets by shuffling the mapping between sets of images and their corresponding leaf labels (as well as hypernyms, in a non-zero hypernym ablation). 
Our first counterfactual dataset, \textbf{Across-category shuffle}, is created by randomly shuffling sets of images and leaves in a manner that destroys visual coherence---e.g., \textit{crow} is mapped to images of \textit{kayak}, \textit{cardinal} is mapped to images of \textit{hummus}, etc., such that members of birds are mapped to visual representations of items that belong to completely different categories.
Our second counterfactual dataset, \textbf{Within-category shuffle}, is created by randomly shuffling the image-leaf pairings \textit{within} categories. In this dataset, the same set of images represent the bird category as they do in reality, but their labels are counterfactual to the real world. 
For example, \textit{crow} is mapped to images of \textit{penguin}, \textit{cardinal} is mapped to images of \textit{eagle}, etc. 
In this way, the internal visual coherence within each category is preserved. 
\Cref{fig:counterfactual-data} shows examples of each type of shuffle, in comparison to the original configuration, and \Cref{appsec:shuffling} shows descriptions of the algorithms used to implement these shuffles.

To quantify visual coherence, we extract DINOv2 embeddings for all training images using mean pooling over image patch tokens, then calculate average pairwise cosine similarities of image embeddings within each category---the higher this value, the higher the visual coherence of that category. We find the average visual coherence of the original and across-category shuffle to be 0.27 (\textit{SD}=.07) and 0.12 (\textit{SD}=.02), respectively (the coherence for within-category shuffle is identical to that of the original by definition). This corroborates our intuition about reduced coherence in the across-category shuffle, relative to the original and within-category configurations.

In both of these counterfactual shuffles, we preserve the language-based lexical-semantic relationships, and manipulate only the visual coherence of the categories---the category \textit{bird} still contains the same lexical items (\textit{crow, cardinal, parrot,} etc.). 
This experiment is analogous in spirit to recent work testing the extent to which LMs learn ``impossible'' languages \citep{kallini-etal-2024-mission, xu2025can}---e.g., LMs struggle to learn languages whose sequences are created by randomly shuffling sequences from actual human languages. 
Analogously, we shuffle data in a manner that reduces the internal visual similarities of categories---e.g., the bird category contains images of \textit{kayaks}, \textit{hummus}, \textit{bandages}, etc. This makes it difficult to conceive of any conceptual structure along which these would cohere \citep{murphy1985role}. 

\begin{figure}[!t]
    \centering
    \includegraphics[width=\linewidth]{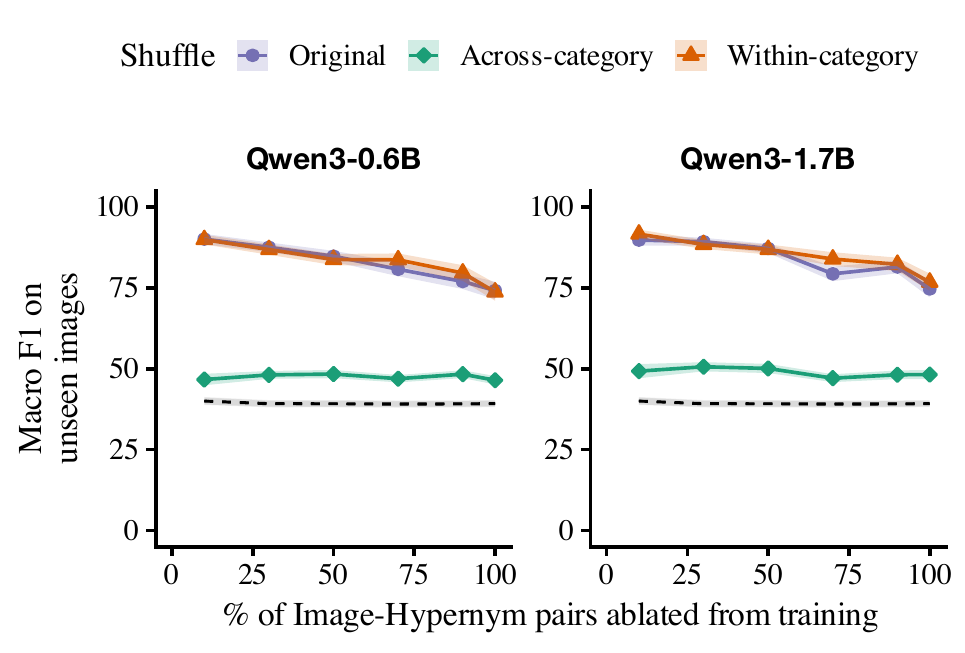}
    \caption{Macro F1 of models trained on data with different counterfactual shuffles (or lack thereof) at predicting unseen hypernyms, across various amounts of exposure to hypernyms. Dashed line indicates macro F1 of the majority-label baseline. Error bands represent 95\% confidence intervals across three random seeds and 53 hypernym categories.}
    \label{fig:counterfactual-results}
    \vspace{-1.5em}
\end{figure}

If the LM is simply executing an abstract rule of \texttt{IF crow THEN bird}, without showing any sensitivity to the coherence among its input representations, then we should see no difference in its taxonomic generalization \textit{regardless} of the type of shuffle. 
In this case, both \textbf{Among} and \textbf{Within} category shuffles should show similar performance to the original experiment. 
On the other hand, if LMs \textit{are} sensitive to the coherence among categories in their inputs, then we should see models trained under the \textbf{Within-category shuffle} show similar performance to the model trained on the original data, while those trained on the \textbf{Across-category shuffle} data should be much worse.

\Cref{fig:counterfactual-results} shows our results, specifically for the Random Hypernym Ablation sub-experiment. 
We see that for both LM backbones, models whose projector was trained on data under the \textbf{Across-category shuffle} substantially underperform models trained on data under the \textbf{Within-category shuffle}, which is instead nearly indistinguishable from models trained with the original configuration. These results also hold for the Llama 3.2 models, as shown in \Cref{fig:counterfactual-llama} in \Cref{appsec:llama}. We also see similar results with Qwen3-8B as the backbone (at 100\% ablation, see \Cref{tab:qwen8b} in \Cref{app:qwen3-8b}), suggesting that arbitrary taxonomic reasoning does not emerge even in a larger model.
Together these results lend support to the hypothesis that cross-modal taxonomic generalization is facilitated when the input representations of category members cohere in a systematic way, relative to the hypothesis that LMs use their learned knowledge of lexical relations in an arbitrary, rule-based manner. 

\section{Post-hoc Analyses}

\begin{figure}[!t]
    \centering
    \includegraphics[width=0.6\linewidth]{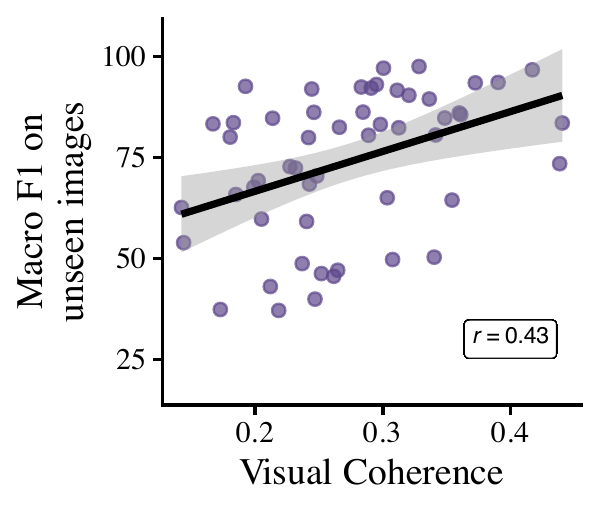}
    \caption{Macro F1 scores on unseen images vs.~Visual coherence across the 53 hypernym categories for the Qwen3-1.7B backbone (at 100\% ablation). $r$ indicates Pearson's correlation.}
    \label{fig:coherence-correlation}
    \vspace{-1em}
\end{figure}

\paragraph{Hypernym-specific results} 
Are our models more likely to show cross-modal generalization for some hypernyms than others? There are two possibilities that can shed light on this question: 1) the LM backbones have stronger hypernymy knowledge for certain hypernyms; and 2) certain hypernyms are more visually coherent. Note that these two are not mutually exclusive (nor are they exhaustive)---it is possible that both have \textit{some} impact on models' hypernym-specific performance. We tease apart their effects by conducting linear mixed-effect model analyses on results from the 100\% ablation subset of the original experiment. Specifically, we predict the models' macro F1 scores using the LMs' hypernymy F1 scores and the visual coherence of the hypernym as fixed effects, and the LM backbone (including different seeds) and the hypernym as random effects. 

We find a significant positive effect of visual coherence ($t$ = 3.338, $p <$ .01), but no effect of the LM backbone F1 scores ($t$ = 0.047, $p$ = .962). \Cref{fig:coherence-correlation} shows an example of the positive relation between visual coherence and cross-modal generalization performance. Overall, these results suggest that cross-modal generalization is greater for hypernyms that are more visually coherent.

\paragraph{Shuffling entire hypernyms}
To further analyze the effect of visual coherence, we conduct a toy experiment on 100\% hypernym ablation in which shuffling is applied at the hypernym level. 
Specifically, we swap leaf labels across a pair of hypernyms so that, for instance, the members of \textit{bird} are now paired with images of \textit{vegetables}. 
In this way, the visual coherence of a given hypernym is still the same, but they are now derived from the images of a different category.
The different number of leaves across hypernyms makes a full hypernym-level swap infeasible; thus we focus on hypernyms with at least 20 leaves and uniformly sample 20 from each (which reduces the data scale and thus limits direct comparability with our shuffling experiments in \cref{sec:counterfactual}).
We then select 5 pairs of hypernyms whose average visual coherence scores differ by less than 0.02 (dessert--furniture, vehicle--vegetable, bird--school supply, clothing accessory--medical equipment, weapon--container).
Finally, we randomly exchange the leaf labels within each selected pair.
For fair comparison, we run an analogous experiment using the same data but without the hypernym-level shuffle. 

Both models achieve generally high F1 scores on the leaf-level data---93.8 and 95.2, respectively, for the unshuffled and the shuffled experiments. 
On held-out hypernyms, the models obtain F1 scores of 57.5 and 51.5, respectively, for the unshuffled and shuffled cases, both of which are above chance (here, 34.7 F1). 
A t-test reveals no significant difference between the unshuffled and shuffled experiments' results ($t$ = 1.30, $p$ = .19). 
Although these F1 scores are not as high as in our original shuffling experiments, shuffling at the hypernym level (while maintaining visual coherence) results in similar performance as in the unshuffled case, suggesting a non-zero effect of visual coherence.

\section{Discussion}
Our results point to two primary takeaways. 
First, given exposure only to lower level categories in a new modality (here, vision), LMs can generalize to higher-level categories. 
That is, they show cross-modal taxonomic generalization. 
Second, the emergence of this generalization depends on the extent to which the mapping between images and lower-level categories is visually coherent.
That is, models' cross-modal taxonomic generalization is not realized in an arbitrary manner, and instead is sensitive to the coherence of their input signal.

These findings reinforce and extend our understanding of how meaning obtained from relational grounding in language \citep{abdou-etal-2021-language, patel2022mapping, mollo2023vector} interacts with those that are obtained from more perceptual signals. Past work in this space has largely shown that models demonstrate reasoning about perceptually relevant concepts by only providing them as text input.
For instance, \citet{patel2022mapping} showed that LMs like GPT-3 \citep{brown2020language} generalize concepts like spatial relations, cardinal directions, and colors to counterfactual worlds (represented as text) by encoding \textit{how} concepts are related. 
That is, LMs could predict which direction was \textit{south}, given partial access to cardinal directions even in rotated worlds, presumably because they encoded the relationship between \textit{north} and \textit{south} (unchanged by rotation), indicating strong relational grounding. 

At first glance, our second takeaway may seem to conflict with these results---after all, if models encode that \textit{bird} and \textit{animal} are hypernyms of \textit{crow}, then they should generalize to these hypernyms regardless of the counterfactual condition. 
However, a closer comparison between our counterfactual data creation and that of \citeauthor{patel2022mapping} suggests that the two conclusions are in fact compatible. 
Specifically, the counterfactual mappings of \citeauthor{patel2022mapping} are \textit{structure-preserving}---it is not the case that the opposite of \textit{east} is \textit{south} in their counterfactual worlds, and therefore the ways in which directions \textit{cohere} are not changed. In comparison, our \textbf{Across-category shuffle} is \textit{not} structure preserving---e.g., \textit{crow} is mapped to images of \textit{kayaks}, and \textit{cardinal} is mapped to images of \textit{hummus}, which means that the members of the bird category are far apart within the input representational space (i.e., visual features). 
The \textbf{Within-category shuffle}, on the other hand, \textit{is} structure preserving. That is, the set of image representations of the members of a given category is unchanged---only the labels are shuffled.
We found models trained under this counterfactual scenario to show similar performance as those trained on the original data. 

Overall, our results lend preliminary support to a more general hypothesis about the importance of \textit{input coherence} in facilitating cross-modal generalization.~This is \textit{preliminary} because our current conception of coherence is based on image features, whereas what makes categories coherent has been posited to rely on more general conceptual knowledge and intuitive theories of the world \citep{murphy1985role, gopnik1994theory, kemp2009structured, murphy2004big}.
Nonetheless, we expect our methodology to inspire a thorough exploration of this connection between categorical coherence and generalization across modalities and phenomena in the future.

\section*{Limitations}

\paragraph{Lack of naturalistic VLM training} 
We use a simple visual question answering-like task for our experiments, one that explicitly focuses on the phenomenon of interest. 
While this allows us to isolate model behavior, VLMs are typically trained on multiple tasks, including but not limited to image captioning \citep{liu2023visual, Qwen2-VL, yang2025qwen3, chen2023pali3visionlanguagemodels, deitke2025molmo}. 
This can raise questions about the implications of our findings for practical VLM training.
At the same time, it would have been infeasible to repeatedly train models on large image captioning datasets, considering the large number of training runs in this work.

\paragraph{Single pair of modalities and single phenomenon} 
Another way in which our results and takeaways are restricted is that our experiments are limited to a single pair of modalities and focus on one phenomenon (hypernymy). 
In our preliminary experiments, we tried to investigate generalization to spatial relations---e.g., can models learn to answer questions about the relation \textit{above} without ever seeing it explicitly in training, presumably by using \textit{below} as indirect evidence? 
However, our models failed to satisfy the important precondition of learning the task in the first place, corroborating recent evidence about the brittle performance of VLMs on spatial reasoning tasks \citep{kamath-etal-2023-whats, chen2025why, zhang2025do}. 
This made it impossible to test for any generalization, since there was no evidence that the model had learned anything in the first place. We expect that as visual encoders and VLMs become more capable in general, our methodology will be useful for studying broader claims about cross-modal generalization to other domains \citep{lewis2019distributional, lupyan2020effects, liu2025learning, wang2026constructing}.

\paragraph{Model scale}  
Our experiments are limited to models of certain (modest) sizes.  It is possible that different generalization behaviors emerge at different model sizes.  Studying the scaling behavior of cross-modal generalization would be an interesting direction for future work.

\paragraph{Single language} 
Finally, our stimuli are limited to the English language. Extensions of this work to multi-lingual settings would lend further evidence about the role of relational grounding from language. 
One particularly exciting extension could be to train for leaf-level information on one language and test for hypernymy knowledge in another---i.e., to test for cross-modal cross-lingual taxonomic generalization.

\section*{Acknowledgments}

We are grateful to Jessica Yang, Zhewei Sun, Andrew Lampinen, Martin Zettersten, Gary Lupyan, Phillip Isola, and Ellie Pavlick for helpful suggestions and questions. We also thank the three anonymous reviewers for their feedback. Kanishka Misra is supported by the Donald D. Harrington Faculty Fellowship at UT Austin.


\bibliography{alternate}

\appendix

\section{Training Details}
\label{appsec:training-details}

We use the code implementation of VLM training from \citet{karamcheti2024prismatic}, as it allows for flexible changes to the image encoder and LM backbone components. 
We adopt the hyperparameters in \Cref{tab:hyp} for all our experiments. 
Model training is conducted on a GPU cluster using 4 RTX A6000/RTX 6000 Ada/L40S/RTX Pro 6000 GPUs, depending on availability.

\begin{table}[h]
\centering
\begin{tabular}{lr}
\toprule
\textbf{Hyperparameter} & \textbf{Value} \\
\midrule
Epochs & 5 \\
Batch Size & 16 \\
Number of Layers & 2 \\
Input Dimension & 1024/1152 \\
Output Dimension & 1024/2048 \\
Max Gradient Norm & 1.0 \\
Weight Decay & 0.1 \\
Learning Rate & $2 \times 10^{-5}$ \\
Activation Function & GeLU \\
Optimizer & AdamW \\
Scheduler & Warmup \& Cosine Decay \\
Warmup Ratio & 0.03 \\
\bottomrule
\end{tabular}
\caption{Training hyperparameters used in our experiments. We adopt a $2$-layer multi-layer perceptron architecture with GeLU activation for all vision-language projectors. The input dimension of our trained MLP projectors is 1024 for DINOv2 and 1152 for SigLIP. 
The output dimension of our trained MLP projectors is 1024 for Qwen3-0.6B, and 2048 for all other LM backbones. The two MLP layers share the same output dimension across experiments.}
\label{tab:hyp}
\end{table}

\section{Details of Experimental Data}
\label{appsec:dataset-details}

\subsection{Data Use and Release Policy} We use images from the THINGS database for our experimental data \citep{things2019}. These images were released with the Attribution CC BY license.\footnote{See \url{https://osf.io/jum2f/files/52wrx}} In our release, we point to their file names and do not release their images. We release our data alongside our code on github,\footnote{\url{https://github.com/sally-xu-42/cross-modal-taxonomic-gen}} which uses an MIT License.

\subsection{Data Statistics}

\begin{figure}
    \centering
    \includegraphics[width=.8\linewidth]{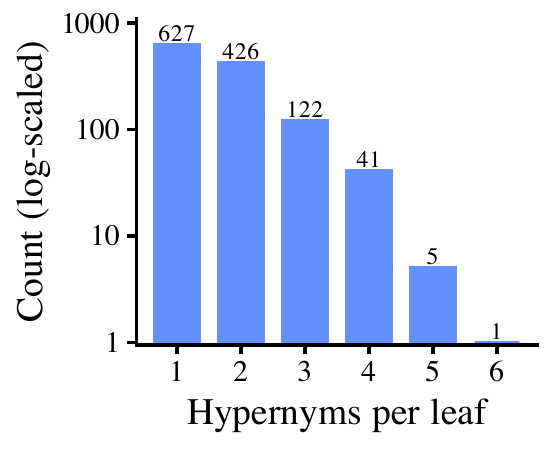}
    \caption{Distribution of the number of hypernym categories per leaf in our dataset. Counts (y-axis) are log-scaled.}
    \label{fig:hyp-per-leaf}
\end{figure}

\begin{figure*}
    \centering
    \includegraphics[width=\linewidth]{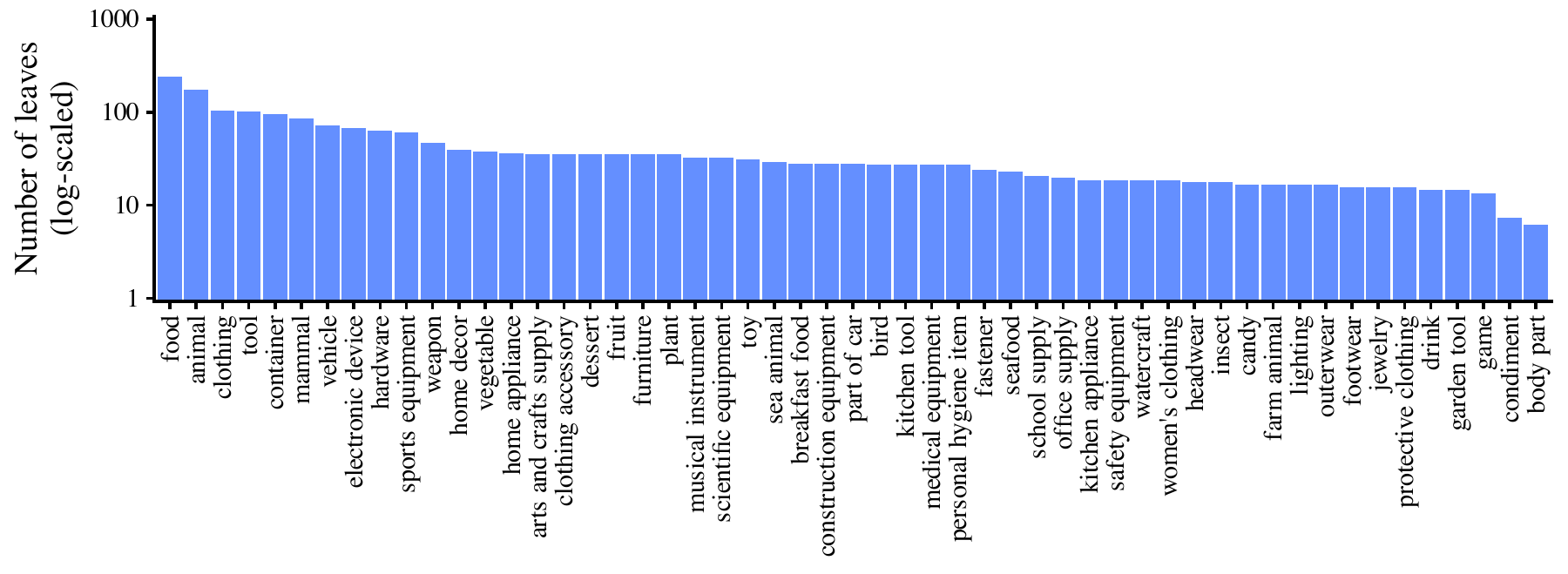}
    \caption{Number of leaves per hypernym category. Counts (y-axis) are log-scaled.}
    \label{fig:hypernym-distribution}
\end{figure*}

\Cref{fig:hyp-per-leaf} shows the distribution of the number of hypernyms per leaf, and \Cref{fig:hypernym-distribution} shows the number of leaves across all 53 hypernyms in the dataset.

\Cref{tab:stats-concept} reports the dataset statistics under five levels of random hypernym ablation and three data variants.
For each ablation level, we show the number of image-question pairs in the training, validation, and test sets, with validation and test sets further divided into leaf-level, seen-hypernym, and unseen-hypernym subsets.
The \textit{original} variant corresponds to the unshuffled dataset.
The \textit{across-category} and \textit{within-category} variants are obtained by shuffling hypernym annotations across categories or within the same category, with details listed in \Cref{sec:counterfactual} and \Cref{appsec:shuffling}.
Across all ablation levels and data variants, the images in the test set are held fixed.

\begin{table*}[!ht]
\centering
\resizebox{\textwidth}{!}{%
\begin{tabular}{@{}rrrrrrrrrr@{}}
\toprule
\multirow{2}{*}{\textbf{\begin{tabular}[c]{@{}c@{}}Ablation\\ level (\%)\end{tabular}}} & \multirow{2}{*}{\textbf{\begin{tabular}[c]{@{}c@{}}\# Leaves\\ affected\end{tabular}}} & \multirow{2}{*}{\textbf{Data variant}} & \multirow{2}{*}{\textbf{\begin{tabular}[c]{@{}c@{}}Train set\\(\# I-Q pairs)\end{tabular}}} & \multicolumn{3}{c}{\textbf{Validation set (\# I-Q pairs)}} & \multicolumn{3}{c}{\textbf{Test set (\# I-Q pairs)}} \\ \cmidrule(l){5-10} 
 &  &  &  & \textbf{leaf} & \textbf{seen} & \textbf{unseen} & \textbf{leaf} & \textbf{seen} & \textbf{unseen} \\ \midrule
\multirow{3}{*}{100} & \multirow{3}{*}{1,222} & Original & 23,250 & 2,028 & -- & 3,380 & 9,394 & -- & 15,702 \\
 &  & Across-category & 23,250 & 2,028 & -- & 3,366 & 9,394 & -- & 15,680 \\
 &  & Within-category & 23,250 & 2,028 & -- & 3,380 & 9,394 & -- & 15,664 \\ \midrule
\multirow{3}{*}{90} & \multirow{3}{*}{1,148} & Original & 24,840 & 2,028 & 134 & 3,246 & 9,394 & 644 & 15,058 \\
 &  & Across-category & 24,818 & 2,028 & 142 & 3,224 & 9,394 & 630 & 15,050 \\
 &  & Within-category & 24,890 & 2,028 & 142 & 3,238 & 9,394 & 660 & 15,004 \\ \midrule
\multirow{3}{*}{70} & \multirow{3}{*}{976} & Original & 28,962 & 2,028 & 492 & 2,888 & 9,394 & 2,300 & 13,402 \\
 &  & Across-category & 28,908 & 2,028 & 488 & 2,878 & 9,394 & 2,278 & 13,402 \\
 &  & Within-category & 28,952 & 2,028 & 496 & 2,884 & 9,394 & 2,308 & 13,356 \\ \midrule
\multirow{3}{*}{50} & \multirow{3}{*}{768} & Original & 34,914 & 2,028 & 1,020 & 2,360 & 9,394 & 4,708 & 10,994 \\
 &  & Across-category & 34,712 & 2,028 & 1,028 & 2,338 & 9,394 & 4,626 & 11,054 \\
 &  & Within-category & 34,972 & 2,028 & 1,004 & 2,376 & 9,394 & 4,754 & 10,910 \\ \midrule
\multirow{3}{*}{30} & \multirow{3}{*}{513} & Original & 43,340 & 2,028 & 1,750 & 1,630 & 9,394 & 8,134 & 7,568 \\
 &  & Across-category & 43,410 & 2,028 & 1,768 & 1,598 & 9,394 & 8,142 & 7,538 \\
 &  & Within-category & 43,354 & 2,028 & 1,680 & 1,700 & 9,394 & 8,158 & 7,506 \\ \midrule
\multirow{3}{*}{10} & \multirow{3}{*}{170} & Original & 55,270 & 2,028 & 2,760 & 620 & 9,394 & 12,948 & 2,754 \\
 &  & Across-category & 55,074 & 2,028 & 2,782 & 584 & 9,394 & 12,850 & 2,830 \\
 &  & Within-category & 55,186 & 2,028 & 2,776 & 604 & 9,394 & 12,894 & 2,770 \\ \bottomrule
\end{tabular}%
}
\caption{Data statistics under $5$ random hypernym ablation levels and counterfactual data variants. ``I-Q pair'' refers to an image and associated yes-no question.  ``Seen'' and ``Unseen'' indicate the number of seen and unseen hypernym I-Q pairs, respectively. ``\# Leaves affected'' is the number of leaf categories that have had any I-Q pairs removed,
accounting for duplicates (removal under multiple hypernyms), and rounding.}
\label{tab:stats-concept}
\end{table*}

\Cref{tab:stats-cat} reports dataset statistics under systematic hypernym ablation at four ablation levels.
In contrast to random ablation, systematic hypernym ablation removes entire hypernym categories from the training data.

\begin{table*}[!ht]
\centering
\resizebox{\textwidth}{!}{%
\begin{tabular}{lrrrrrrrr}
\toprule
\multirow{2}{*}{\textbf{\begin{tabular}[l]{@{}l@{}}Ablation level\\\small{(hypernyms removed)}\end{tabular}}} &
\multirow{2}{*}{\textbf{Data variant}} &
\multirow{2}{*}{\textbf{Train set}} &
\multicolumn{3}{c}{\textbf{Validation set}} &
\multicolumn{3}{c}{\textbf{Test set}} \\
\cmidrule(lr){4-6}\cmidrule(lr){7-9}
& & & \textbf{leaf} & \textbf{seen} & \textbf{unseen} & \textbf{leaf} & \textbf{seen} & \textbf{unseen} \\
\midrule

\multirow{3}{*}{full}
  & Original        & 23,250 & 2,028 & --   & 3,380 & 9,394 & --    & 15,702 \\
  & Across-category        & 23,250 & 2,028 & --   & 3,366 & 9,394 & --    & 15,680 \\
  & Within-category  & 23,250 & 2,028 & --   & 3,366 & 9,394 & --    & 15,664 \\
\midrule

\multirow{3}{*}{40}{}
  & Original        & 29,892 & 2,028 & 564  & 2,816 & 9,394 & 2,684   & 13,018 \\
  & Across-category        & 30,014 & 2,028 & 600  & 2,766 & 9,384 & 2,722   & 12,958 \\
  & Within-category  & 29,890 & 2,028 & 548  & 2,832 & 9,384 & 2,694   & 12,970 \\
\midrule

\multirow{3}{*}{30}{}
  & Original        & 33,800 & 2,028 & 924 & 2,456 & 9,394 & 4,242 & 11,460 \\
  & Across-category        & 33,702 & 2,028 & 888 & 2,478 & 9,394 & 4,242 & 11,438 \\
  & Within-category  & 33,844 & 2,028 & 926 & 2,454 & 9,394 & 4,254 & 11,410 \\
\midrule

\multirow{3}{*}{20}{}
  & Original        & 40,730 & 2,028 & 1,516 & 1,864 & 9,394 & 7,076 & 8,626 \\
  & Across-category        & 40,804 & 2,028 & 1,522 & 1,844 & 9,394 & 7,068  & 8,612 \\
  & Within-category  & 40,708 & 2,028 & 1,528 & 1,852 & 9,394 & 7,066  & 8,598 \\
\midrule

\multirow{3}{*}{10}{}
  & Original        & 48,214 & 2,028 & 2,142 & 1,238 & 9,394 & 10,094 & 5,608 \\
  & Across-category        & 48,120 & 2,028 & 2,176 & 1,190 & 9,394 & 10,028 & 5,652 \\
  & Within-category  & 48,230 & 2,028 & 2,136 & 1,244 & 9,394 & 10,090 & 5,574 \\

\bottomrule
\end{tabular}%
}
\caption{Data statistics under $4$ systematic hypernym ablation levels and counterfactual data variants. ``Seen'' and ``Unseen'' indicate the number of seen and unseen hypernyms, respectively.}
\label{tab:stats-cat}
\end{table*}

\Cref{tab:ablated-categories} lists the specific hypernym categories removed at each systematic ablation level.
These category lists are used consistently across all experiments.

\begin{table*}[!ht]
\centering
\resizebox{\textwidth}{!}{%
\begin{tabular}{cp{15cm}}
\toprule
\textbf{Ablation level} & \textbf{Ablated categories} \\
\midrule

40 &
kitchen appliance, home decor item, medical equipment, plant,
protective clothing item, farm animal, mammal, car part, dessert,
breakfast food, sea animal, hardware item, school supply,
source of light, garden tool, drink, sports equipment, toy,
insect, watercraft, footwear, vegetable, piece of jewelry, game,
condiment, animal, personal hygiene item, bird, musical,
office supply item, item of clothing, piece of women's clothing,
construction equipment, candy, headwear, scientific equipment,
home appliance, seafood, safety equipment, container \\

\midrule
30 &
fastener, condiment, container, kitchen appliance, office supply item,
body part, vegetable, kitchen tool, home appliance, outerwear,
arts and crafts item, construction equipment, game, insect,
breakfast food, home decor item, clothing accessory, food,
school supply, fruit, sports equipment, sea animal,
piece of women's clothing, candy, safety equipment, farm animal,
dessert, scientific equipment, bird, tool \\

\midrule
20 &
medical equipment, sea animal, home decor item, fastener, condiment,
breakfast food, hardware item, office supply item, toy, seafood,
musical, mammal, vehicle, personal hygiene item,
protective clothing item, sports equipment, construction equipment,
insect, scientific equipment, tool \\

\midrule
10 &
container, personal hygiene item, dessert, item of clothing,
furniture, breakfast food, scientific equipment, garden tool,
car part, sea animal \\

\bottomrule
\end{tabular}
}
\caption{List of ablated categories under different systematic hypernym ablation levels.}
\label{tab:ablated-categories}
\end{table*}

\section{Shuffling Algorithms}
\label{appsec:shuffling}

This appendix provides detailed descriptions of the algorithms used to create the counterfactual datasets.

\subsection{Across-category Shuffle}
\label{app:across-category-shuffle}

Algorithm~\ref{alg:across-category-shuffle} describes the procedure for creating the across-category shuffle, which randomly maps original leaf-level concepts to new leaf-level concepts, regardless of their hypernym categories.

\begin{algorithm}[h]
\caption{Across-category Shuffle}
\label{alg:across-category-shuffle}
\KwIn{$\mathcal{H}$: hypernym-to-leaf mapping}
\KwOut{$\mathcal{M}$: concept-to-concept shuffle mapping}
\SetKwFunction{ShuffleConcepts}{ShuffleConcepts}
\SetKwProg{Fn}{Function}{:}{}
\Fn{\ShuffleConcepts{$\mathcal{H}$}}{
    $\mathcal{C} \leftarrow \emptyset$\;
    \ForEach{leaf list $\mathcal{L} \in \mathrm{values}(\mathcal{H})$}{
        $\mathcal{C} \leftarrow \mathcal{C} \cup \mathcal{L}$\;
    }
    $\mathcal{C}_{\text{shuffled}} \leftarrow \text{RandomShuffle}(\mathcal{C})$\;
    $\mathcal{M} \leftarrow \emptyset$\;
    \For{$i = 1$ \KwTo $|\mathcal{C}|$}{
        $\mathcal{M}[\mathcal{C}[i]] \leftarrow \mathcal{C}_{\text{shuffled}}[i]$\;
    }
    \Return{$\mathcal{M}$}\;
}
\end{algorithm}

\subsection{Within-category Shuffle}
\label{app:within-category-shuffle}

Algorithm~\ref{alg:within-category-shuffle} describes the procedure for creating the within-category shuffle, which reassigns concepts to images within the same category while ensuring no concept maps to itself.

\begin{algorithm}[h]
\caption{Within-category Shuffle}
\label{alg:within-category-shuffle}
\KwIn{$\mathcal{H}$: hypernym-to-concepts mapping}
\KwOut{$\mathcal{M}$: concept-to-concept shuffle mapping}
\SetKwFunction{ShuffleWithinCategory}{ShuffleWithinCategory}
\SetKwProg{Fn}{Function}{:}{}
\Fn{\ShuffleWithinCategory{$\mathcal{H}$}}{
    \tcp{Build concept-to-hypernyms mapping}
    $\mathcal{L} \leftarrow \emptyset$\;
    \ForEach{hypernym $h$, concepts $\mathcal{C}_h \in \mathcal{H}$}{
        \ForEach{concept $c \in \mathcal{C}_h$}{
            $\mathcal{L}[c] \leftarrow \mathcal{L}[c] \cup \{h\}$\;
        }
    }
    \tcp{Group concepts by their hypernym sets}
    $\mathcal{G} \leftarrow \emptyset$\;
    \ForEach{concept $c$, hypernyms $\mathcal{H}_c \in \mathcal{L}$}{
        $\mathcal{G}[\text{frozenset}(\mathcal{H}_c)] \leftarrow \mathcal{G}[\text{frozenset}(\mathcal{H}_c)] \cup \{c\}$\;
    }
    \tcp{Shuffle within each group}
    $\mathcal{M} \leftarrow \emptyset$\;
    \ForEach{group key $k$, concepts $\mathcal{C}_k \in \mathcal{G}$}{
        $\mathcal{C}_{\text{shuffled}} \leftarrow \text{RandomShuffle}(\mathcal{C}_k)$\;
        \For{$i = 1$ \KwTo $|\mathcal{C}_k|$}{
            $\mathcal{M}[\mathcal{C}_k[i]] \leftarrow \mathcal{C}_{\text{shuffled}}[i]$\;
        }
    }
    \Return{$\mathcal{M}$}\;
}
\end{algorithm}

\section{Results on Llama 3.2 Models}
\label{appsec:llama}

\begin{figure*}
    \centering
    \includegraphics[width=\linewidth]{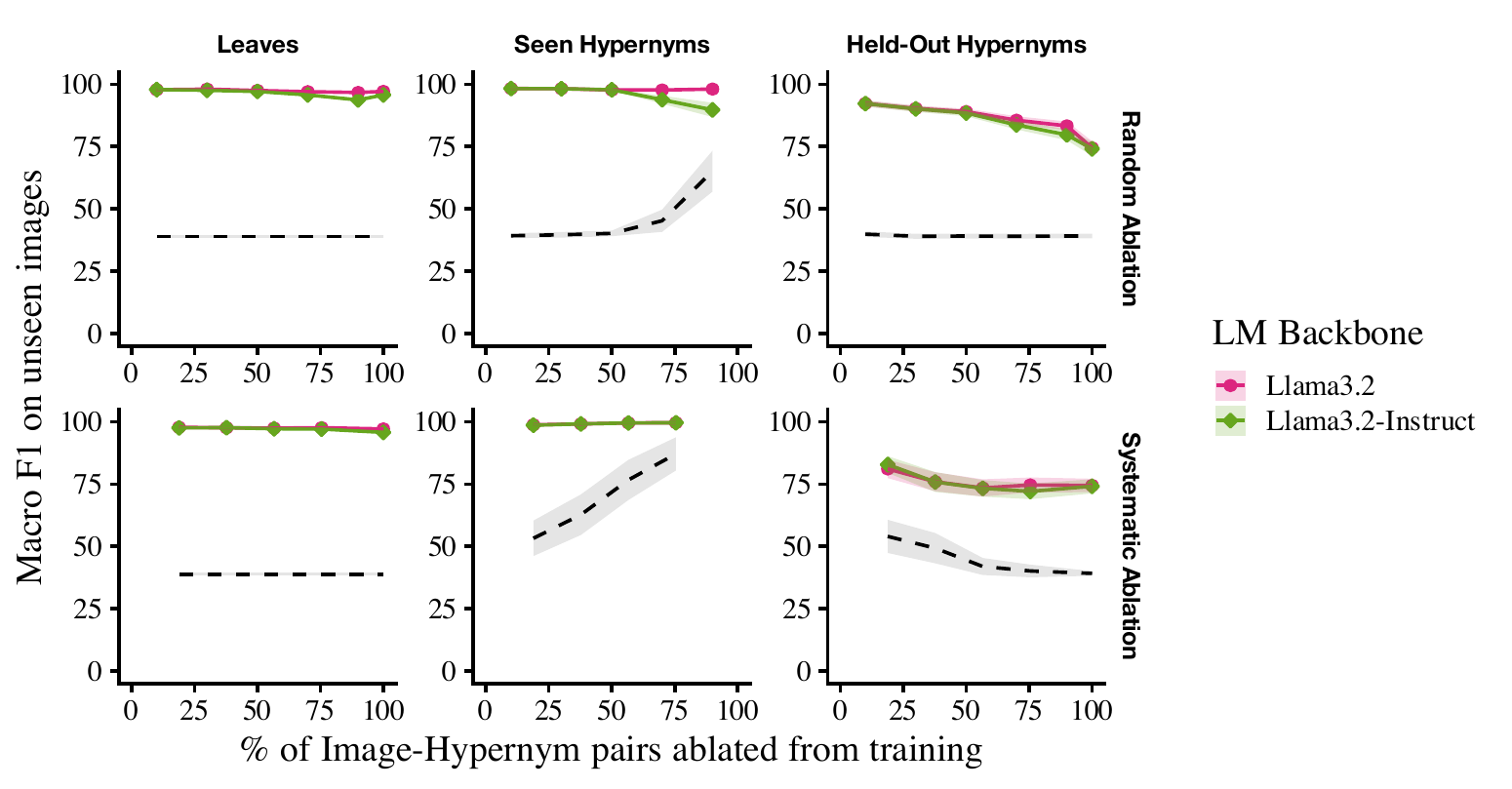}
    \caption{Macro F1s of Llama 3.2 models on unseen images across LM backbones, LM representations (pre-trained vs.~random), and hypernym ablation type (Random vs.~Systematic) at different amounts of exposure to hypernym categories, for various test splits. Dashed line indicates macro F1 of the majority-label baseline. Error bands represent 95\% confidence intervals across random seeds ($N$ = 3) and categories (hypernyms = 53, leaves = 1216). }
    \label{fig:main-exp-llama}
\end{figure*}

\begin{figure*}
    \centering
    \includegraphics[width=0.7\linewidth]{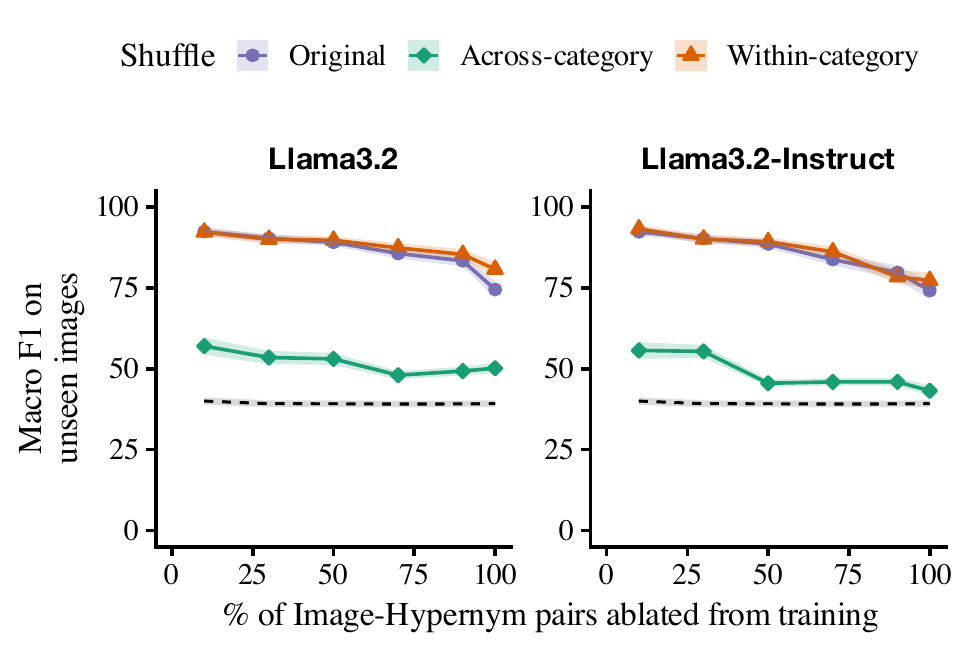}
    \caption{Macro F1s of Llama 3.2 models trained on data with different counterfactual shuffles (or lack thereof) at predicting hypernyms, across various amounts of exposure to hypernym categories. Dashed line indicates macro F1 of the majority-label baseline. Error bands show 95\% confidence intervals for three random seeds and 53 hypernym categories.}
    \label{fig:counterfactual-llama}
\end{figure*}

In this section, we report results from our experiments on Llama-3.2-1B and its instruction-tuned variant Llama-3.2-1B-Instruct \citep{grattafiori2024llama3herdmodels}. \Cref{fig:main-exp-llama} shows results of our main experiment on the Llama 3.2 models, and \Cref{fig:counterfactual-llama} shows results on our counterfactual experiments for these models.

\clearpage
\section{Results on Qwen3-8B LM backbone}
\label{app:qwen3-8b}

\Cref{tab:qwen8b} shows results from the 100\% ablation experiments across various counterfactual shuffles, using the Qwen3-8B model as the LM backbone. We see similar results as with smaller models in the main paper---results on Original and Within-category shuffles are similar while those on Across-category shuffle are vastly different (and lower). This suggests that arbitrary reasoning about hypernymy does not emerge in this larger model, and that it is still sensitive to the visual coherence of categories in the incoming input.

\begin{table}[!t]
    \centering
    \begin{tabular}{lrr}
    \toprule
        \textbf{Experiment Setting} & \textbf{Macro F1} & \textbf{Std. Error} \\ \midrule
        Original & 78.4 & 1.3\\
        Across-category & 50.0 & 0.6\\
        Within-category & 79.8 & 1.2\\ 
        Majority-label & 39.2 & 0.9\\
        \bottomrule
    \end{tabular}
    \caption{Macro F1 of models trained with Qwen3-8B as the LM backbone, on the 100\% ablation setting, across various shuffles (or lack thereof). Standard error calculated over 3 seeds and 53 hypernym categories.}
    \label{tab:qwen8b}
\end{table}

\section{Cross-Modal Generalization at Varying Depths of Taxonomic Organization}
\label{app:depthwise}

\begin{figure}
    \centering
    \includegraphics[width=\linewidth]{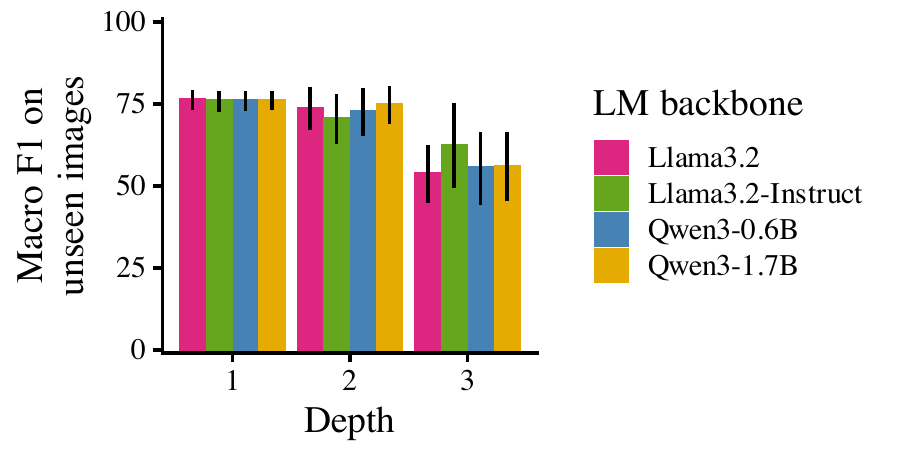}
    \caption{Macro F1s on unseen images for all four LM backbones across various depths of category organization in our analysis, computed for the 100\% ablation experiment. Error bars indicate 95\% confidence intervals across three random seeds and number of categories per depth ($N_1$ = 41; $N_2$ = 9; $N_3$ = 3).}
    \label{fig:depthwise}
\end{figure}

Our set of 53 categories can be organized hierarchically into multiple tree-like structures, each representing a taxonomy of a specific higher level category. The maximum depth across these trees is 3 (e.g., \textit{kitchen tool} $\prec$ \textit{tool} $\prec$ \textit{hardware}, where `$\prec$' represents the hypernym relation, with the right hand side category being the hypernym of the left hand side category). Overall, we have 41 categories at depth 1, 9 categories at depth 2, and 3 categories at depth 3. 
To understand how cross-modal taxonomic generalization of our tested models varies across these depths, we present results (here, at the 100\% ablation level) divided across the three depths for all models in \Cref{fig:depthwise}. We see that models generally perform better at depth 1 and 2 than at depth 3, i.e., they are markedly better at generalizing to lower level categories than at higher level categories.

Note that it is not clear if the poorer performance at depth 3 is necessarily because the models struggle at multi-hop reasoning (over hierarchically organized category structures), since the internal mechanism of how models arrive at their answer is not obvious. An alternate explanation could be the `frequency of instantiation' hypothesis \citep{barsalou1985ideals}, which states that the number of times a learner observes a given category as member of another category determines the strength of the category relationship between the two. Therefore, the LM backbones may have encountered fewer instances of hyponym/leaf concepts as members of depth 3 hypernyms relative to depth 1 and 2 hypernyms. This does not appeal directly to a process that commits to an explicit hierarchical representation of categories, and yet can explain the observed result. We therefore caution readers from making inferences about the nature of explicit multi-hop hierarchical reasoning in the models from these results.

\end{document}